\documentclass[sigconf]{acmart} 

\renewcommand\footnotetextcopyrightpermission[1]{} 
\usepackage{multirow, makecell}			
\usepackage{booktabs}					
\usepackage{tabularx}					
\usepackage{threeparttable}				
\usepackage{siunitx}
\usepackage{tabularx}
\AtBeginDocument{%
  \providecommand\BibTeX{{%
    \normalfont B\kern-0.5em{\scshape i\kern-0.25em b}\kern-0.8em\TeX}}}

\setcopyright{acmcopyright}
\copyrightyear{2022}
\acmYear{2022}
\acmDOI{}

\acmConference[CBMI '22]{}{}{Graz, Austria}
\acmPrice{15.00}
\acmISBN{978-1-4503-XXXX-X/18/06}

\acmSubmissionID{}

\begin{document}

\title{Hybrid Transformer Network for Deepfake Detection}

\author{Sohail Ahmed Khan}
\email{sohail.khan@uib.no}
\affiliation{%
  \institution{MediaFutures}
  \city{Bergen}
  \country{Norway}
}

\author{Duc-Tien Dang-Nguyen}
\email{ductien.dangnguyen@uib.no}
\affiliation{%
  \institution{MediaFutures}
  \city{Bergen}
  \country{Norway}}


\begin{abstract}
Deepfake media is becoming widespread nowadays because of the easily available tools and mobile apps which can generate realistic looking deepfake videos/images without requiring any technical knowledge. With further advances in this field of technology in the near future, the quantity and quality of deepfake media is also expected to flourish, while making deepfake media a likely new practical tool to spread mis/disinformation. Because of these concerns, the deepfake media detection tools are becoming a necessity. In this study, we propose a novel hybrid transformer network utilizing early feature fusion strategy for deepfake video detection. Our model employs two different CNN networks, i.e., (1) XceptionNet and (2) EfficientNet-B4 as feature extractors. We train both feature extractors along with the transformer in an end-to-end manner on FaceForensics++, DFDC benchmarks. Our model, while having relatively straightforward architecture, achieves comparable results to other more advanced state-of-the-art approaches when evaluated on FaceForensics++ and DFDC benchmarks. Besides this, we also propose novel face cut-out augmentations, as well as random cut-out augmentations. We show that the proposed augmentations improve the detection performance of our model and reduce overfitting. In addition to that, we show that our model is capable of learning from considerably small amount of data.
\end{abstract}

\ccsdesc[500]{Security and privacy~Social aspects of security and privacy}

\keywords{deepfake detection, face forensics, attention mechanisms, image analysis, feature fusion, misinformation detection, transformers}



\maketitle

\section{Introduction}
The availability of huge image/video datasets and affordable compute resources, has resulted in swift progress in the field of deep learning research, specifically in the subarea of Generative Adversarial Networks (GANs) \cite{IGoodfellow}. This progress has made it almost effortless to generate realistic synthetic content even for non-technical computer users. The synthetic content generated using deep learning models (i.e., GANs) is called Deepfake media. Deepfake media can be in the form of images, videos, text and audios. However, out of all the different categories of deepfake media, the visual deepfake media is the most common form of fake/synthetic content we encounter nowadays. The number of deepfake media generation techniques is growing exponentially. The newer generation techniques are able to generate extremely plausible synthetic content, and it is becoming more and more challenging to detect the generated fake media. 


\begin{figure*}[htb]
\hspace*{-0.2cm}
\begin{minipage}[b]{1.0\linewidth}
  \centering
  \centerline{\includegraphics[width=1.0\linewidth, height=0.40\linewidth]{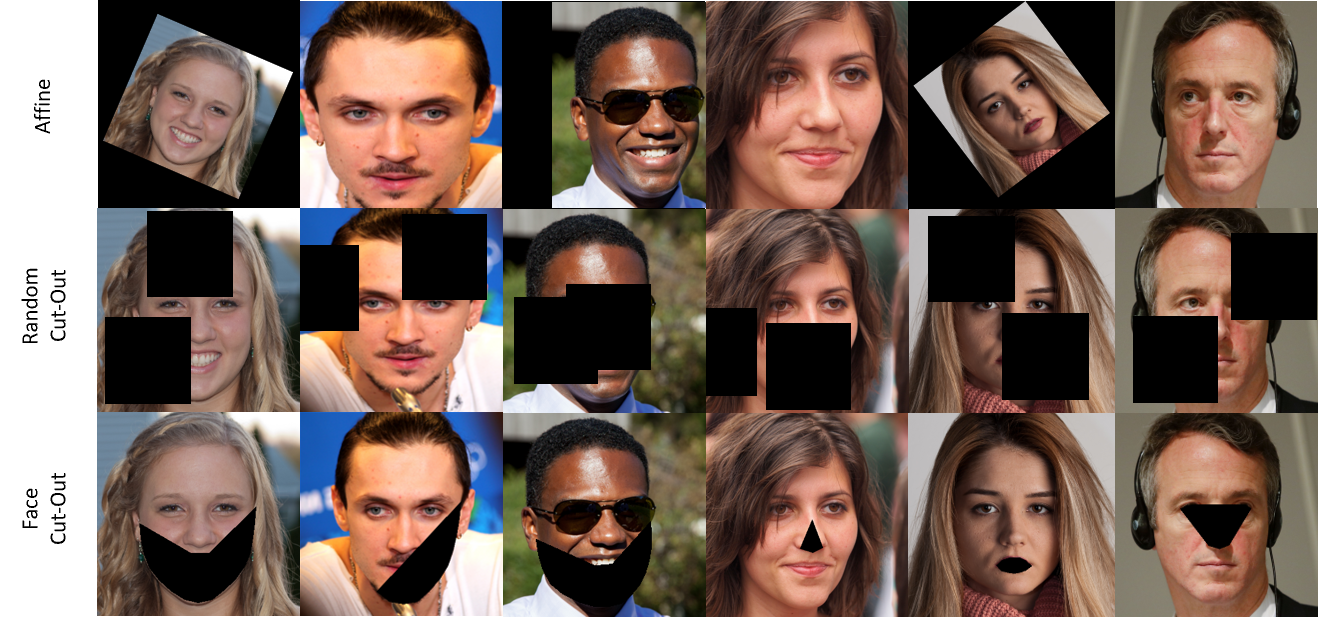}}
\end{minipage}
\caption{Our face pre-processing module is responsible for applying 3 different types of image augmentations, (1) Affine Transformations, (2) Random cut-out augmentations, and (3) Face cut-out augmentations. More details about the augmentations we use are given in upcoming sections. We show that the employed image augmentations improve detection performance while reducing overfitting. Photos acquired from Flickr Faces HQ dataset \cite{Karras2019ASG}.}
\label{fig:res}
\end{figure*}

The most popular form of facial deepfake media we encounter at present is generated using face swapping method, in which the face of a person (target) is swapped with the face of another person (source). There are 4 different types of facial deepfake media, i.e., (1) Face Swapping, (2) Face Re-enactment, (3) Face Editing and (4) Face Synthesis \cite{YMirsky}. In this paper we focus on detecting facial deepfake media, specifically the media generated using face swapping and face re-enactment techniques.

Using ensembled or fusion based models tend to achieve exceptional results as compared to single models \cite{DFDC, Qi2020DeepRhythmED, Zhu2021FaceFD, Dolhansky2020TheDD}. We therefore propose to employ two different CNN models as feature extractors along with a transformer architecture (Vision Transformer \cite{Dosovitskiy2021AnII}). We expect that by fusing features extracted using different feature extractors will result in diverse feature spaces, which will help the transformer to learn diverse set of features. Transformer architectures are capable of simultaneously learning meaningful associations from long input sequences. We therefore choose transformer \cite{Dosovitskiy2021AnII} to learn joint feature space, instead of the classical way of using a fully connected layer to combine different feature sets. Besides this, hybrid (having CNN as feature extractor instead of using simple patch embeddings) transformer models tend to achieve even better results.

We choose XceptionNet as one of the feature extractors as it has been widely employed in deepfake detection domain \cite{Rossler2019FaceForensicsLT}. The second CNN which we choose is the EfficientNet B4 model, which also achieved exceptional results on ImageNet benchmark. We do not freeze the feature extractors during training, i.e., we train both feature extractors, as well as the transformer architecture in an end-to-end manner using a single loss function.


The contributions of this paper are three fold, (1) we propose a novel hybrid transformer architecture which learns from joint feature sets extracted by two different CNN feature extractors, (2) we show that image augmentations we generate using our face pre-processing module, combined with other affine transformation based augmentations, improve the performance of the detection models while reducing overfitting, and (3) we show that while having a simple and easy to implement architecture, our model achieves comparable results to other more complex state-of-the-art approaches while being trained on comparably smaller number of data samples. 

This paper is structured as follows, in section \ref{sec:relatedworks} we present a brief literature review, in section \ref{sec:methodology} we describe the augmentations we employ, the proposed face pre-processing module, our model architecture, the datasets used to train our models, and implementation details, in section \ref{sec:results} we compare the achieved results with other deepfake detection baselines, and in section \ref{sec:conclusion} we conclude our study and propose future research directions.


\section{Related Work}
\label{sec:relatedworks}
Recent works on deepfake media detection mostly employ CNN based architectures along with other strategies (e.g., multimodal features, recurrent networks, transformer models etc) to detect deepfake images/videos. Unlike the previous research studies, in this paper we propose a novel strategy to simultaneously learn from joint feature spaces extracted using two different CNN feature extractors using a transformer architecture while employing heavy image augmentations.

    
    
    
    
    

 Rossler {\textit{et al.}} in \cite{Rossler2019FaceForensicsLT} proposed a simple deepfake detection technique based on the XceptionNet \cite{xception} CNN model pre-trained on the imagenet dataset. Authors fine-tune the generic XceptionNet on their FaceForensics++ dataset while reporting excellent performance scores the four subsets of the FaceForensics++ dataset, namely,  (1) FaceSwap, (2) Face2Face, (3) DeepFakes, and (4) NeuralTextures \cite{Rossler2019FaceForensicsLT}. The proposed model achieved excellent results on uncompressed videos, however, lost performance when tested on compressed videos.
 
 In \cite{Zhu2021FaceFD} Zhu \emph{et al.} propose to utilize 3D facial details to detect deepfakes. Authors find that merging the 3D identity texture and direct light is significantly helpful in detecting deepfakes. They employ the XceptionNet CNN model for feature extraction. A face cropped image and its 3D detail is used to train the detection model. They also perform a detailed analysis of a number of different feature fusion strategies. The proposed technique was trained on FaceForensics++ dataset and evaluated on (1) FaceForensics++, (2) Google Deepfake Detection Dataset, and (3) DFDC datasets. Authors report promising results on all of the three datasets along with better generalization capability than the previously proposed techniques. 
 
Qi \emph{et al.} in \cite{Qi2020DeepRhythmED} propose a novel deepfake detection technique which they call, DeepRhythm. The proposed technique works by analyzing the heartbeat rhythms of a person in a given video. They employ photoplethysmography (PPG) to analyze minute changes in the skin tone inherent with the blood pumping visible on the human faces. Authors evaluate the proposed technique on FaceForensics++, and DFDC datasets and report excellent performance results.

In \cite{XXuan} Xuan \emph{et al.} proposed a more general deepfake media detection technique by employing image augmentations, for example, gaussian blur and gaussian noise to preprocess images in order to remove low level high frequency GAN artifacts present inside the generated images. They then trained a forensic convolutional neural network model on the preporcessed images. They established that by using image augmentations on both real and fake images, destroy the low
level noise cues, while forcing the forensics model to learn
more meaningful features. Through experimentation authors establish the effectiveness of their proposed technique on deepfake media detection.

Sabir \emph{et al.} in \cite{ESabir} proposed a recurrent convolutional network for deepfake media detection. The DenseNet convolutional neural network was employed along with a gated recurrent neural network to exploit temporal inconsistencies present between frames of a deepfake video. The proposed technique was evaluated on the FaceForensics++ \cite{Rossler2019FaceForensicsLT} dataset showing promising performance.

\begin{figure*}[!htb]
\hspace*{-0.2cm}
\begin{minipage}[b]{1.0\linewidth}
  \centering
  \centerline{\includegraphics[width=1.0\linewidth, height=0.45\linewidth]{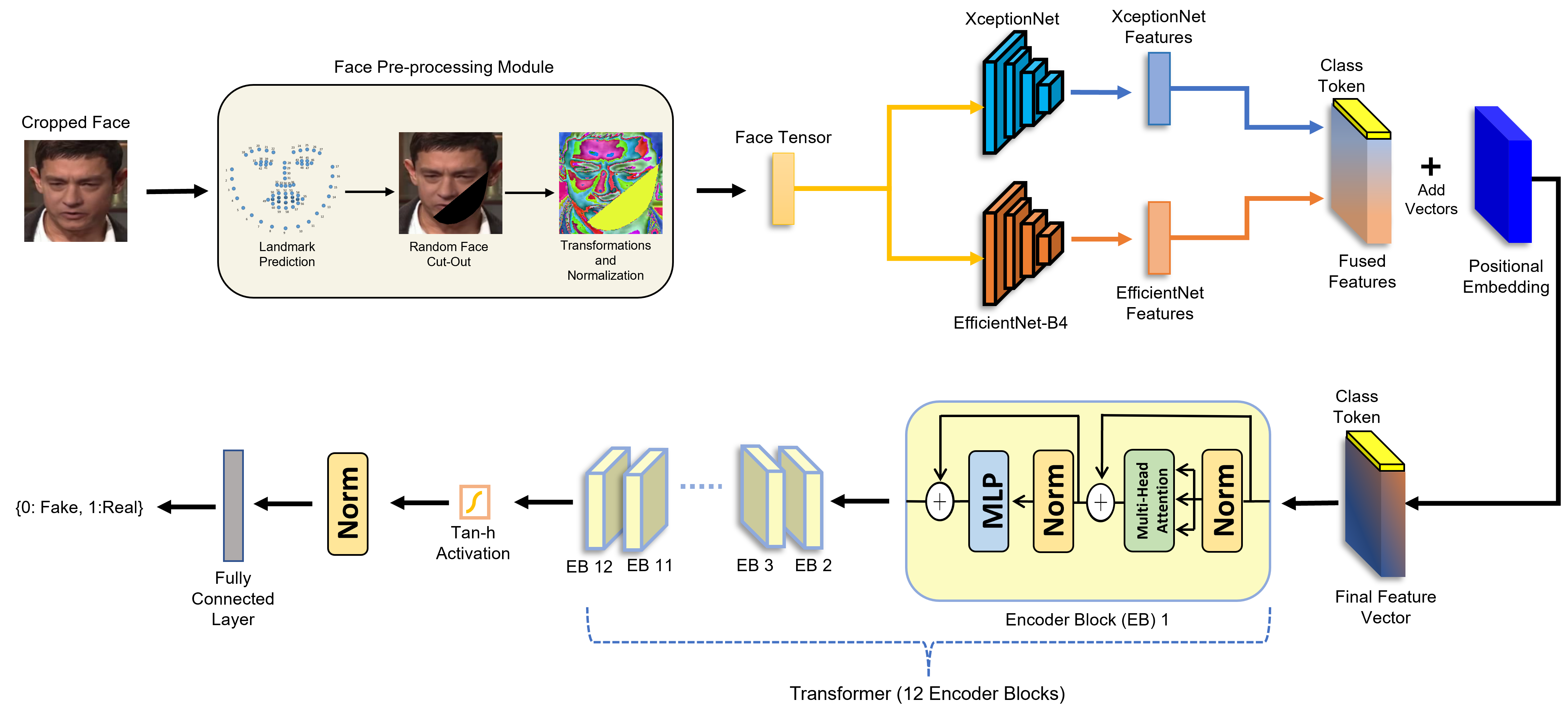}}
\end{minipage}
\caption{Our hybrid transformer network comprising of two different CNNs, (1) XceptionNet, (2) EfficientNet-B4 and a transformer. Our model is trained on the cropped face images. The face image is fed to the face pre-processing module which applies face cut-out/random cut-out and affine transformation augmentations randomly. The augmented face image is then fed to the CNN feature extractors. The extracted features are then fused together by concatenation. A BERT style $[class]$ token and learnable positional embedding are added to the concatenated features which are then fed to the transformer. Both feature extractors and the transformer are trained in an end-to-end manner using a single loss function.}
\label{fig:res}
\end{figure*}

Following a similar path, Guera and Delp in \cite{DGuera} proposed to employ a convolutional neural network along with a long short term memory (LSTM) network for deepfake video detection. Authors tried to exploit inter-frame discrepancies inherent to most deepfake videos. The CNN was tasked with extracting frame-level features, which were then fed to the LSTM network to learn the temporal features. Authors evaluated the model on their own dataset showing exceptional performance.

Nguyen \emph{et al.} in \cite{HNguyen} proposed a capsule network based model to detect deepfake media. The proposed model was evaluated on four different deepfake detection datasets containing a wide range of synthetic images and videos. The proposed method achieved excellent results as compared to other methods on all datasets.

Again in \cite{HNguyen2}, Nguyen \emph{et al.} proposed a different strategy employing a Y-shaped encoder-decoder model. Authors trained the model by following a multi-task learning based technique which was able to classify and generate a segmentation mask of the tampered regions within manipulated images/videos. The proposed model was evaluated on the FaceForensics and FaceForensics++ \cite{Rossler2019FaceForensicsLT} datasets achieving promising results even when finetuned on a small number of images.

In \cite{UACiftci} Ciftci \emph{et al.} proposed a novel deepfake detection technique employing biological signals (i.e., photoplethysmography or PPG signals) to train a CNN and a support vector machine (SVM). The predictions from the CNN and SVM were fused to get final classification label. The proposed model achieved promising results when evaluated on several different deepfake detection datasets including, Face Forensics, Face Forensics++ \cite{Rossler2019FaceForensicsLT}, and CelebDF \cite{Li2020CelebDFAL} datasets.

In \cite{DAfchar} Afchar \emph{et al.} proposed two different CNN models, which they called, (1) Meso-4 and (2) MesoInception-4, both containing a small number of layers focusing on mesoscopic image features. The proposed model was tested on an existing deepfake detection dataset, as well as, a custom dataset collected by the authors. The model achieved excellent results on both datasets.

In \cite{Khan2021VideoTF} proposed a novel video transformer network for deepfake media detection, capable of learning from new data in an incremental manner. The proposed video transformer model was trained on multimodal data (i.e., face cropped images, and UV texture maps). The proposed models achieved excellent results on a number of different deepfake detection benchmarks.

\section{Methodology}
\label{sec:methodology}
In this section, we briefly introduce, (1) image augmentations we employ to train our models, (2) face pre-processing module which is responsible for applying the image aumentations and normalizing the image, (3) the two CNN backbones we employ to extract image features, and (4) the transformer model.

\subsection{Augmentations}
We use two different image cropping (face cut-out, random cut-out) based augmentations, which are applied along with three different affine transformation (rotation, translation, scaling) based augmentations. 

In face cut-out augmentations, we crop out specific face part from the image in a random order using facial landmarks, as shown in figure. Whereas, in case of random cut-out augmentations, we crop out random square shaped region from a given face image. Both of these augmentations are separately applied to train two different models. However, the affine transformation based augmentations are applied along with both of these cut-out based augmentations. The pipeline shown in figure \ref{fig:res}, applies face cut-out augmentation along with affine transformation based augmentations to a given face image.

\subsection{Face Pre-processing Module}
The face pre-processing module is responsible for applying augmentations and normalizing a given input face image. In case of face cut-out augmentations, the face pre-processing module takes a face cropped image as input, predicts 81 face landmarks using OpenCV's facial landmarks predictor. It then selects a number of different landmarks randomly to cut a specific face part out from the image. 

The face pre-processing module in case of random cut-out augmentations, takes a face cropped image, randomly crop outs two squared shaped regions from the image. The purpose of using heavy cut out augmentations is that we want to prevent our model from overfitting. If we do not use augmentations, the model will memorize the training data and will not be able generalize well on the test set, as can be seen from the results in table \ref{tab:aug_no_aug}.

\subsection{XceptionNet}
XceptionNet architecture with depth-wise separable convolutions was proposed by François Chollet in \cite{xception}. For deepfake detection, Rossler \emph{et al.} employed XceptionNet in \cite{Rossler2019FaceForensicsLT}. They showed that the XceptionNet architecture achieved exceptional results on FaceForensics++ dataset. Because of the excellent performance of XceptionNet on deepfake detection, in this paper we use XceptionNet as one of the feature extractors.

\subsection{EfficientNet-B4}
EfficientNets are a new set of state-of-the-art convolutional neural network models for image classification proposed in \cite{Tan2019EfficientNetRM}. For deepfake detection EfficientNet architectures achieved promising results on DFDC dataset. In-fact, the winning model of the DFDC was an ensemble of 5 EfficientNet CNNs \cite{DFDC}. Furthermore, because of the memory constraints we needed a smaller but effective model, making EfficientNet-B4 the best choice for our study.

\subsection{Hybrid Image Transformer}

\begin{table*}[!htb]
  \centering
    \caption{Performance (accuracy) comparison of a number of different deepfake detection baseline models on FaceForensics++ dataset. Each of the mentioned model was trained on all subsets of the FaceForensics++ dataset at once. Best results are highlighted.}
    \begin{tabular}{l|c|c|c|c|c|c}
    \toprule
    \multicolumn{1}{c|}{\textbf{Approach}} & \textbf{Deepfakes} & \textbf{Face2Face}& \textbf{FaceSwap} & \textbf{NeuralTextures} & \textbf{Pristine} & \textbf{Cumulative}\\
    \midrule
    \midrule
    Steg. Features + SVM \cite{Fridrich2012RichMF} & 68.80\% & 67.69\%& 70.12\%& 69.21\%& 72.98\% & 70.97\%\\
    Cozzolino \emph{et al.} \cite{Cozzolino2017RecastingRL} & 75.51\% & 86.34\%& 76.81\%& 75.34\%& 78.41\% & 78.45\%\\
    Bayar and Stamm \cite{Bayar2016ADL} & 90.25\%& 93.96\%& 87.74\%& 83.69\%& 77.02\% & 82.97\%\\
    Afchar \emph{et al.} \cite{DAfchar} & 89.55\%& 88.60\%& 81.24\%& 76.62\%& 82.19\% & 83.10\%\\
    Rossler \emph{et al.} \cite{Rossler2019FaceForensicsLT} & 97.49\%& 97.69\%& 96.79\%& \textbf{92.19\%}& 95.41\% & 95.73\%\\
    Qi \emph{et al.} \cite{Qi2020DeepRhythmED} &\textbf{ 99.70\%}& \textbf{98.90\%}& 97.80\%& - & - & - \\
    Ours (Face cut-out) & 97.85\%& 97.85\% & 96.42\% & 90.71\%& 95.00\% & 95.57\%\\
    Ours (Random cut-out) & 98.57\%& 98.57\% & \textbf{97.85\%}& 92.14\%& \textbf{97.85\%} & \textbf{97.00\%}\\
    \bottomrule
    \end{tabular}%
  \label{tab:comparison_stats}%
\end{table*}%

\begin{table*}[!ht]
  \centering
    \caption{Performance comparison of our models trained with and without image augmentations. In this table, DF refers to Deepfakes, F2F refers to Face2Face, FS refers to FaceSwap, NT refers to NeuralTextures, and P refers to Pristine.}
    \begin{tabular}{l|c|c|c|c|c|c}
    \toprule
    \multicolumn{1}{c|}{\textbf{Model}} & \textbf{DF} & \textbf{F2F} & \textbf{FS} & \textbf{NT} & \textbf{P} & \textbf{Agg.}\\
    \midrule
    \midrule
    No Augs & 95.71\% & 93.57\% & 92.85\% &  85.00\%& 96.42\% & 92.71\%\\
    Face cut-out  & 97.85\%& 97.85\% & 96.42\% & 90.71\%& 95.00\% & 95.57\% \\
    \textbf{Random cut-out} & \textbf{98.57\%}& \textbf{98.57\%}& \textbf{97.85\%}& \textbf{92.14\%}& \textbf{97.85\%} & \textbf{97.00\%}\\
    \bottomrule
    \end{tabular}%
  \label{tab:aug_no_aug}%
\end{table*}%

Table \ref{tab:aug_no_aug} presents a performance comparison of our model trained under different augmentation strategies. We use heavy image augmentations for example, rotation, horizontal flipping, translation, scaling, face cut-outs and random cut-outs. During experimentation we found that the model overfits severely when no image augmentations are employed. Random cut-out augmentations along with affine transformation based augmentations give best results.

\begin{table}[]
  \centering
    \caption{Number of frames used to train models on different datasets: Pristine, FaceSwap, Deepfakes, Face2Face, Neural Textures.}
    \begin{tabular}{l|c|c|c}
    \toprule
    \multicolumn{1}{c|}{\textbf{Dataset}} & \textbf{Training} & \textbf{Validation} & \textbf{Test} \\
    \midrule
    \midrule
    Pristine & 100K & 20K & 2.24K \\
    FaceSwap & 25K & 5K & 2.24K \\
    Deepfakes & 25K & 5K & 2.24K \\
    Face2Face & 25K  & 5K   & 2.24k \\
    Neural Textures & 25K  & 5K   & 2.24K \\
    \bottomrule
    \end{tabular}%
  \label{tab:training_data_stats}%
\end{table}%

\begin{table*}[]
  \centering
    \caption{We compare the performance (accuracy) of our model with the XceptionNet proposed by Rossler \emph{et al.} in \cite{Rossler2019FaceForensicsLT}. In this table, DF refers to Deepfakes, F2F refers to Face2Face, FS refers to FaceSwap, NT refers to NeuralTextures.}
    \begin{tabular}{l|c|c|c|c|c}
    \toprule
    \multicolumn{1}{c|}{\textbf{Model}} & \textbf{DF} & \textbf{F2F} & \textbf{FS} & \textbf{NT} & \textbf{Agg.}\\
    \midrule
    \midrule
    Rossler \emph{et al.} \cite{Rossler2019FaceForensicsLT} & 92.48\% & 91.33\% & 92.63\% & 85.98\% & 90.60\%\\
    \textbf{Ours (Random cut-out)} &  \textbf{95.00\%}&  \textbf{95.00\%}&  \textbf{95.71\%}&  \textbf{90.00\%}& \textbf{93.92\%}\\
    \bottomrule
    \end{tabular}%
  \label{tab:small_train}%
\end{table*}%

\begin{table}[!htb]
  \centering
    \caption{Number of training and validation images used by different deepfake detection techniques. The number of train, validation, and test sets of other studies in this table are rough estimates, as the the authors do not specify exact number of images they used to train their models.}
    \begin{tabular}{l|c|c|c}
    \toprule
    \multicolumn{1}{c|}{\textbf{Approach}} & \textbf{Train} & \textbf{Validation} & \textbf{Test}\\
    \midrule
    \midrule
    Rossler \emph{et al.} \cite{Rossler2019FaceForensicsLT} & 388K & 70K & 70K\\
    Zhu \emph{et al.} \cite{Zhu2021FaceFD} & 360K & 70K & 70K\\
    Ours & 200K & 40K & 11.2K\\
    \bottomrule
    \end{tabular}%

  \label{tab:train_size_comparison}%
\end{table}%

We employ BERT \cite{Devlin2019BERTPO, Dosovitskiy2021AnII} styled transformer to learn the joint features extracted using the two CNNs. Since Transformers have proven to be excellent in simultaneously learning meaningful properties from long sequences because of their bidirectional representation learning capability, we employ these models to learn the joint features extracted by two CNNs. We believe that rather than utilizing the early feature fusion followed by a fully connected layer to learn from the joint feature space, as it is widely done while using multiple CNNs; using a transformer to learn the joint feature space will yield better results. 

In addition to this, ensembled/fusion based models are proven to achieve better results when compared with single models specifically in deepfake detection, as we can infer from Facebook's Deepfake Detection Challenge, in which most of the top ranked models employed ensembled/fusion based networks \cite{DFDC}.

We utilize Transformer architecture pretrained on ImageNet\footnote{https://github.com/lukemelas/PyTorch-Pretrained-ViT}) on features extracted using an early fusion based strategy in which two different CNN models as introduced above. The extracted features are concatenated, a classification token (similar to BERT's \textit{[class]} token) is added at the start, and assigned a positional embedding before being fed to the transformer architecture. Our model comprises of 12 encoder blocks and 12 attention heads (following the architecture of ViT-Base-16 \cite{Dosovitskiy2021AnII}). The weights for the pretrained CNN models are obtained from Ross Wightman\footnote{https://github.com/rwightman/pytorch-image-models}. Both the feature extractors, and the transformer are trained in an end-to-end manner using a single loss function i.e., we do not freeze the weights of feature extractors while training.

\subsection{Dataset}
We train and evaluate our models on FaceForensics++ \cite{Rossler2019FaceForensicsLT} and DFDC \cite{Dolhansky2020TheDD} datasets. 

\subsubsection{\textbf{FaceForensics++}}
This dataset comprises of four subsets namely, (1) FaceSwap, (2) Face2Face, (3) Deepfakes, and (4) Neural Textures. The videos we used to train our models are high quality (c23). We create train, validation, and test sets according to the instructions in FaceForensics++ dataset paper \cite{Rossler2019FaceForensicsLT}. There are 720 train videos, 140 validation videos, and 140 test videos. We train our models on 200K images (100k real and 100k fake), which is less than half of the size of training set Rossler \emph{et al.} used to train their XceptionNet model in \cite{Rossler2019FaceForensicsLT}. Out of the 200K training images, 160K images are used for training and the remaining 40K images are used for validation.

For fake videos, we extract 50 face frames starting from the beginning of each video, whereas, for the real videos, we extract 150 frames from each video. We do this to balance the real and fake test sets. The exact amount of images from each dataset used to train, validate and test our models as given in table \ref{tab:training_data_stats}. 

To evaluate our models we extract 16 face frames from 140 test videos which results in a total of 2100 images as mentioned in the table \ref{tab:training_data_stats}. To assign a classification label to any test video, we take 16 frames and feed to our model one by one. The final prediction is made after averaging the predictions obtained from the model for each frame.

\subsubsection{\textbf{Deepfake Detection Challenge (DFDC)}} This dataset comprises of around 124K videos. To train our models we use only around 8K videos. The amount of fake videos in DFDC dataset is more than the real videos, and thus to balance the real and fake image samples, we extract 50 frames from each fake video, whereas, from each real video we extract 150 frames. This results in around 265K real and fake frames, from which we only use 48K images to train our model and 12K images for validation purposes. So in total, we only used 60K real and fake face frames from DFDC dataset to train and validate our models.

We use 400 test videos provided with the DFDC dataset to evaluate our model. For evaluation, we use the same strategy we used for evaluating the models on FaceForensics++ dataset, i.e., we extract 16 face frames from each test video and feed to our model one by one. The final prediction about a video is made after averaging the individual frame predictions.

\subsection{Implementation Details}
For face detection and cropping, we use OpenCV\footnote{https://opencv.org/}. For custom image augmentations such as, rotation, flipping, translation and cutouts we employ ImgAug\footnote{https://imgaug.readthedocs.io/en/latest/} library. We use Ross Wightman's github repoitory to download CNN weights pretrained on ImageNet. We borrow code and weights of vision transformer pretrained on ImageNet from Luke Melas \footnote{https://github.com/lukemelas/PyTorch-Pretrained-ViT}. 

To train our models we use SGD with a momentum ranging from 0.6 to 0.9, with a learning rate of \(3\times10^{-3}\). We stop training when the validation loss keeps on increasing for 3 consecutive epochs, or the training accuracy approaches to 100\% (to prevent severe overfitting). We train the two CNNs and the transformer in an end-to-end manner, and optimize them through a single binary cross entropy loss function.

We resize images to $[3, 224, 224]$ dimensions. Having higher resolution images yield better results but because of memory constraints we choose to use this image resolution to train and evaluate our models. The input image is fed to the CNN feature extractors, which after extracting features, each of the feature extractors return features of dimension $[1, 162, 768]$. The obtained features are then concatenated to get final features of dimension $[1, 324, 768]$. We append a BERT style $[class]$ token at the start of the extracted features resulting making the dimension $[1, 325, 768]$. A learnable positional embedding is added to these features through element wise addition. The resulting features are then fed to the transformer.

\section{Results}
\label{sec:results}
In this section we will present and compare the results our model achieved on the FaceForensics++ and DFDC dataset. We trained our model on FaceForensics++ dataset under three different settings e.g., (1) without image augmentations, (2) with face cut-out augmentations, and (3) with random cut-out augmentations. We found that the model trained using random cut-out augmentations outperformed the other two variants, and thus we further trained this model on DFDC dataset as well. Performance comparison of our model under three augmentation strategies can be seen in table \ref{tab:aug_no_aug}. As we understand, the reason behind the excellent performance of random cut-out augmentation is because it cuts out most parts of the image in a random order preventing the model from memorizing face images. 

It should also be noted that we trained our models on smaller number of samples from the FaceForensics++ dataset (given in table \ref{tab:training_data_stats}) as compared to other approaches, e.g., Rossler \emph{et al.} \cite{Rossler2019FaceForensicsLT} train their model on around $388K$ images, Zhu \emph{et al.} \cite{Zhu2021FaceFD} train their models on around $360K$ images. We compare our models with the baseline results provided in the original FaceForensics++ paper \cite{Rossler2019FaceForensicsLT}, since they provide performance scores of each model on every subset of the FaceForensics++ dataset. We also present the results achieved by our models on DFDC dataset in table \ref{tab:dfdc_comparison_stats}. 

Table \ref{tab:comparison_stats} presents a detailed comparison of the obtained accuracy scores on all of the subsets of FaceForensics++ dataset. Our model trained using heavy image augmentations achieves comparable results to the baseline techniques listed in Table \ref{tab:comparison_stats} and new state-of-the-art deepfake detection techniques\cite{Zhu2021FaceFD, Qi2020DeepRhythmED}. The reason for choosing the techniques in table \ref{tab:comparison_stats} for comparison is that these techniques are also evaluated in the original FaceForensics++ dataset paper, and carry out the training in the same manner as we do in this paper. Testing is done in a different manner than these approaches i.e., we test our models on 16 face frames per video, whereas these approaches test there models on 100 face frames per video. For validation, the approaches in FaceForensics++ paper use 100 face frames per video, whereas, we split the train and validation set in 80:20 ratio i.e., 200K images for training and 40K images for validation.

Some of the other deepfake detection state-of-the-art techniques \cite{Zhu2021FaceFD, Qi2020DeepRhythmED, UACiftci} achieve better results than our model, however, those techniques are quite complex to implement, and in most cases use more data than we use to train the models.

In table \ref{tab:small_train} we present a comparison of our model with XceptionNet \cite{Rossler2019FaceForensicsLT} model proposed by Rossler \emph{et al}. We train and evaluate our model on each of the four subsets of FaceForensics++ dataset separately. For training we only use 5000 fake images (from each subset) and 5000 real images. While Rossler \emph{et al.} trained their models on 50 videos or 13500 images since they use 270 images from each video for training. We show that our model even being trained on less data, improves detection performance. In table \ref{tab:train_size_comparison} we compare the size of training set Rossler \emph{et al.} \cite{Rossler2019FaceForensicsLT} and Zhu \emph{et al.} \cite{Zhu2021FaceFD} use to train their respective models.

\begin{table}[!htb]
  \centering
    \caption{Performance (accuracy) comparison of a number of different deepfake detection baseline models on DFDC dataset. Best results are highlighted.}
    \begin{tabular}{l|c|c|c}
    \toprule
    \multicolumn{1}{c|}{\textbf{Approach}} & \textbf{Dataset} & \textbf{Train Size} & \textbf{Accuracy}\\
    \midrule
    \midrule
    Mittal \emph{et al.} \cite{TMittal} & DFDC & - & 84.40\%\\
    Wodajo \emph{et al.} \cite{Wodajo2021DeepfakeVD} & DFDC & 112K & 91.50\%\\
    Bondi \emph{et al.} \cite{Bondi2020TrainingSA} & DFDC & - & 92.20\%\\
    Ours (Random cut-out) & DFDC & 48K & \textbf{98.24\%}\\
    \bottomrule
    \end{tabular}%

  \label{tab:dfdc_comparison_stats}%
\end{table}%

We present a comparison of results our model achieved on the DFDC dataset in table \ref{tab:dfdc_comparison_stats}. We show that while being trained on smaller training set, our model still achieves exceptional performance scores as compared to other relevant works proposed in the past. 

\section{Conclusion and Future Work}
\label{sec:conclusion}
Detecting deepfake media is crucial as well as challenging. Besides other challenges of deepfake media detection, for example, poor generalization capability of the detection models, deepfake media is also adversarial in nature and continues to evolve rapidly. In this study we presented an early fusion based hybrid transformer network for deepfake media detection. Our model achieved comparable results to most of the state-of-the-art deepfake detection techniques \cite{Zhu2021FaceFD, Qi2020DeepRhythmED} . After this, we plan to train and evaluate our model on other prominent deepfake detection datasets, such as, Celeb-DF \cite{Li2020CelebDFAL}, ForgeryNet \cite{He2021ForgeryNetAV} and others. We also plan to analyze the generalization capability of our model on unseen data in future studies, while trying to visually interpret our model to know what kind of features it utilizes more while making decisions and how it differentiates between different categories of deepfake media, i.e., face swapping, face re-enactment etc.

In future we will focus our work mainly on improving the generalization capability of the deepfake detection models, and further work on improving the adversarial robustness of the detection models. 

\section{Acknowledgments}
This research was supported by industry partners and the Research Council of Norway with funding to MediaFutures: Research Centre for Responsible Media Technology and Innovation, through the Centres for Research-based Innovation scheme, project number 309339.
\bibliographystyle{ACM-Reference-Format}
\bibliography{main}

\end{document}